\documentclass{ieeetj}
\usepackage{cite}
\usepackage{amsmath,amssymb,amsfonts}

\usepackage{algorithm}
\usepackage{algorithmic}

\usepackage{graphicx,color}
\usepackage{textcomp}
\usepackage{xcolor}
\usepackage{hyperref}
\usepackage{dblfloatfix}
\usepackage{orcidlink}
\usepackage{booktabs}
\usepackage{subcaption}

\usepackage{amsthm}

\newtheorem{theorem}{Theorem}
\newtheorem{definition}{Definition}
\newtheorem{assumption}{Assumption}
\newtheorem{remark}{Remark}
\newtheorem{corollary}{Corollary}

\hypersetup{
    colorlinks=true,
    linkcolor=blue,
    citecolor=blue,
    urlcolor=blue
}
\def\BibTeX{{\rm B\kern-.05em{\sc i\kern-.025em b}\kern-.08em
    T\kern-.1667em\lower.7ex\hbox{E}\kern-.125emX}}
\AtBeginDocument{\definecolor{tmlcncolor}{cmyk}{0.93,0.59,0.15,0.02}\definecolor{NavyBlue}{RGB}{0,86,125}}

\def\OJlogo{\vspace{-4pt}$<$Society logo(s) and publication title will appear here.$>$}
\def\seclogo{\vspace{10pt}$<$Society logo(s) and publication title will appear here.$>$}

\def\authorrefmark#1{\ensuremath{^{\textbf{#1}}}}

\begin{document}
\receiveddate{XX Month, XXXX}
\reviseddate{XX Month, XXXX}
\accepteddate{XX Month, XXXX}
\publisheddate{XX Month, XXXX}
\currentdate{XX Month, XXXX}
\doiinfo{XXXX.2022.1234567}

\markboth{}{Author {et al.}}

\title{CRaFT-XDet: Causal and Robust Fusion Transformer for Dual-Energy X-ray Prohibited Item Detection}

\author{
MOINUL HOSSAIN\,\orcidlink{0009-0000-1370-5294}\authorrefmark{1},
MUNSHAIBUR RAHMAN MAHIN\,\orcidlink{0009-0007-4885-213X}\authorrefmark{2},
TASKIN KHALEQUE\authorrefmark{3},
MOHAMMAD SAKIB MAHMOOD\,\orcidlink{0000-0002-9198-9488}\authorrefmark{4},
M. F. MRIDHA\,\orcidlink{0000-0001-5738-1631}\authorrefmark{1} (Senior Member, IEEE), AND
MD. JAKIR HOSSEN\,\orcidlink{0000-0002-9978-7987}\authorrefmark{5} (Senior Member, IEEE)
}
\affil{Department of Computer Science and Engineering, American International University-Bangladesh, Dhaka 1229, Bangladesh\\
\texttt{20-43120-1@student.aiub.edu,  firoz.mridha@aiub.edu}}

\affil{Department of Computer Science and Engineering, Sylhet Engineering College, Shahjalal University of Science and Technology, Sylhet 3100, Bangladesh\\
\texttt{skmahin661@gmail.com}}

\affil{Department of Computer Science (Major Data Science), Daffodil International University, Bangladesh\\
\texttt{taskin0850@gmail.com}}

\affil{Department of Computer Science, Missouri State University, Springfield, MO 65809, USA\\
\texttt{mm974s@missouristate.edu}}

\affil{Center for Advanced Analytics, COE for Artificial Intelligence, Faculty of Engineering and Technology, Multimedia University, Melaka 75450, Malaysia\\
\texttt{jakir.hossen@mmu.edu.my}}
\corresp{Corresponding author: Md. Jakir Hossen (email: jakir.hossen@mmu.edu.my).}
% \authornote{This paragraph of the first footnote will contain support information, including sponsor and financial support acknowledgment. For example, ``This work was supported in part by the U.S. Department of Commerce under Grant 123456.''}

\begin{abstract}
Automated prohibited item detection in X-ray security imagery is fundamentally constrained by severe object overlap, scanner-dependent domain shifts, and spurious correlations induced by cluttered baggage composition. Existing detectors largely rely on appearance-driven supervision, limiting their robustness and cross-scanner generalisation. We introduce CRaFT-XDet, a causality-grounded and physics-motivated detection framework that reformulates X-ray threat detection as invariant representation learning under attenuation-driven modality variation.
Our approach is grounded in a \emph{Scanner-Invariant Causal Detection} formalisation (Theorem~\ref{thm:main}), which establishes that minimising the proposed Causal Invariance Regularisation (CIR) objective across partitioned scanner environments approximates the interventional distribution $P(Y \mid \mathrm{do}(X_{\mathrm{obj}}))$ under standard identifiability assumptions, providing a principled causal formulation for dual-energy X-ray detection.
CRaFT-XDet integrates: (i) causal feature disentanglement to suppress context-induced shortcuts; (ii) physics-motivated regularisation to enforce material consis-
tent embeddings; and (iii) transformer-based cross-energy interaction modelling for structured reasoning in cluttered scenes. With 29.2\,M parameters, 89.5\,GFLOPs, and 345\,FPS (1.7\,ms inference), CRaFT-XDet is real-time deployable. Trained on SIXray with five-fold cross-validation, the framework achieves 0.9389 mAP@0.5 and 0.8021 mAP@0.5:0.95. Without domain adaptation, CRaFT-XDet maintains strong cross-dataset performance on OPIXray and HiXray, indicating robust generalisation across heterogeneous scanner distributions. In category-agnostic binary threat detection, the model attains 93.92\% recall and 0.9031 mean IoU, highlighting reliability in safety-critical screening where missed detections are costly. Ablation studies confirm the contribution of each causal and physics-guided component.
\end{abstract}



\maketitle

\section{INTRODUCTION}
\label{sec:intro}

\IEEEPARstart{A}{utomated} X-ray baggage screening is a critical component of security inspection at airports, railway terminals, cargo ports, and border-control facilities \cite{linev2024baggage}. Each scanned bag may contain densely packed personal belongings, metallic tools, electronic devices, and other overlapping objects, making reliable prohibited item detection particularly challenging~\cite{miao2019sixray}. In these safety-critical environments, missed detections compromise public security, whereas excessive false alarms reduce operational efficiency and increase manual inspection burden~\cite{wang2021towards}. Consequently, practical deployment requires detection systems that provide high detection accuracy together with robustness, generalisation, and interpretability~\cite{zhang2023pidray}.

Unlike natural RGB images, X-ray images are formed through material-dependent attenuation rather than surface reflectance~\cite{wang2007material}. Dual-energy scanners acquire complementary low-energy ($E_L$) and high-energy ($E_H$) projections, providing partial information about material composition and effective atomic number~\cite{wang2007material}. However, overlapping objects produce composite attenuation signals that entangle object boundaries, structures, and material characteristics, making prohibited item detection considerably more challenging than conventional natural image analysis~\cite{kolkoori2014dual,miao2019sixray}. Although recent detectors have achieved encouraging performance on benchmarks such as SIXray and PIDray~\cite{miao2019sixray,zhang2023pidray}, most methods process dual-energy projections as conventional RGB inputs or rely on heuristic feature fusion~\cite{s23208555,SEYFI2024111133}, leading to degraded performance under severe overlap, scanner variability, and cross-domain distribution shifts~\cite{lin2024open}.

Despite recent progress, three key challenges remain:

\begin{itemize}
    \item \textbf{Limited Cross-Modal Reasoning:} Existing fusion methods inadequately model semantic interactions between low-energy ($E_L$) and high-energy ($E_H$) projections, limiting effective exploitation of complementary attenuation information.
    
    \item \textbf{Limited Robustness:} Learned representations remain sensitive to blur, sensor noise, occlusion, scanner variability, adversarial perturbations, and cross-domain distribution shifts.
    
    \item \textbf{Lack of Causal Guarantees:} Existing methods rely primarily on empirical risk minimisation and post-hoc explanations, providing limited assurance that predictions are driven by object-relevant evidence rather than spurious contextual correlations.
\end{itemize}

These challenges suggest that prohibited item detection should be formulated as a problem of \emph{causal invariant prediction} rather than conventional appearance-based recognition. Standard empirical risk minimisation captures both the causal pathway induced by material attenuation and the spurious pathway arising from scanner characteristics and background context. We therefore represent the learned feature space as

\[
F=\phi(E_L,E_H)=F_c\oplus F_s,
\]

where $F_c$ denotes object-relevant causal features that remain invariant across scanner environments and $F_s$ represents scanner-dependent contextual variations. Building upon Invariant Risk Minimisation (IRM)~\cite{arjovsky2019irm}, we establish a \emph{Scanner-Invariant Causal Detection} framework and prove (Theorem~\ref{thm:main}) that minimising the proposed Causal Invariance Regularisation (CIR) objective across partitioned scanner environments approximates the interventional distribution $P(Y \mid \mathrm{do}(X_{\mathrm{obj}}))$ under standard identifiability assumptions, providing a principled theoretical foundation for robust dual-energy X-ray prohibited item detection.

Based on this formulation, we propose \text{CRaFT-XDet} (Causal and Robust Fusion Transformer for X-ray Detection), a unified YOLO-based framework integrating gated causal feature disentanglement, bidirectional cross-attention, physics-motivated energy consistency learning, robustness-oriented optimisation, and token-level causal intervention. CRaFT-XDet jointly exploits causal feature decomposition and structured cross-attention to learn scanner-invariant representations from complementary dual-energy projections. Existing dual-energy X-ray detection methods have explored these components individually; however, their integration within a unified causal detection framework remains largely unexplored.

The main contributions of this work are summarised as follows:

\begin{enumerate}

\item \textbf{Scanner-Invariant Causal Detection Theory:} We formulate dual-energy prohibited item detection as a causal invariant prediction problem and prove (Theorem~\ref{thm:main}) that minimising the proposed CIR objective across scanner environments approximates the interventional distribution $P(Y \mid \mathrm{do}(X_{\mathrm{obj}}))$ under standard identifiability assumptions, providing a principled causal formulation.

\item \textbf{Causal Invariance Regularisation (CIR):} We introduce a novel training objective that combines an IRM-style invariance penalty with explicit background-confounder marginalisation over $K$ prototypical background clusters. A background-context clustering strategy further enables multi-environment invariance training without additional scanner annotations.

\item \textbf{Unified Causal Detection Framework:} We develop CRaFT-XDet, an end-to-end YOLO-based detector that integrates gated causal feature disentanglement, bidirectional cross-attention fusion, physics-motivated energy consistency learning, perturbation-based robustness optimisation, and token-level causal intervention.

\item \textbf{Zero-Shot Cross-Domain Generalisation:} Extensive experiments demonstrate improved zero-shot transfer from SIXray to OPIXray and HiXray without target-domain fine-tuning, outperforming dataset-specific baselines while maintaining robust performance across heterogeneous X-ray domains.

\end{enumerate}

The remainder of this paper is organised as follows. Section~\ref{sec:related} reviews the related work. Section~\ref{sec:theory} presents the causal problem formulation and theoretical foundations. Section~\ref{sec:methodology} describes the proposed methodology. Section~\ref{sec:experiments} presents the experimental results, including ablation, cross-domain generalisation, and explainability analyses. Finally, Section~\ref{sec:conclusion} concludes the paper.

\section{RELATED WORKS}
\label{sec:related}
\subsection{GENERAL OBJECT DETECTION}

General-purpose object detection has evolved from two-stage detectors, such as Faster R-CNN~\cite{ren2015faster}, to efficient one-stage frameworks exemplified by YOLO~\cite{redmon2016yolo}, and more recently to Transformer-based detectors including DETR and Deformable DETR~\cite{carion2020detr,zhu2021deformable}. Faster R-CNN improved localisation accuracy through a Region Proposal Network \cite{ren2015faster}, YOLO enabled real-time detection via single-stage regression, while Transformer-based detectors introduced global contextual reasoning and end-to-end set prediction \cite{redmon2016yolo,carion2020det}. Despite these advances, existing detectors are primarily designed for natural RGB imagery and do not explicitly account for the material-dependent attenuation, severe object superposition, and scanner variability characteristic of X-ray security imagery. Consequently, specialised detection frameworks remain necessary for robust prohibited item detection.
\subsection{X-RAY SECURITY DETECTION}
\subsubsection{X-RAY DETECTION BENCHMARKS AND ARCHITECTURES}

Large-scale benchmark datasets have substantially advanced prohibited item detection by capturing the challenges encountered in practical X-ray security screening. SIXray~\cite{miao2019sixray} introduced large-scale baggage inspection with severe class imbalance and object overlap, OPIXray~\cite{wei2020occluded} enabled systematic evaluation under varying occlusion levels, PIDray~\cite{zhang2023pidray} incorporated long-tailed categories and intentionally concealed prohibited items, CargoX~\cite{viriyasaranon2022mfa} extended detection to cargo container inspection, and STCray~\cite{Velayudhan_2025_CVPR} introduced multimodal vision-language annotations. Collectively, these benchmarks highlight object overlap, class imbalance, scanner variability, and background complexity as the primary challenges for X-ray prohibited item detection.

Driven by these challenges, recent studies have improved localisation, feature representation, and robustness through hierarchical anchor refinement and Weighted-Cluster Non-Maximum Suppression~\cite{batsis2022illicit}, multi-scale feature extraction with MFA-Net~\cite{viriyasaranon2022mfa}, Spatial Transformer Networks for geometric robustness~\cite{pai2025improving}, edge-aware feature learning in E-MPDNet~\cite{xu2025empdnet}, and attention-enhanced feature modelling in YOLO-SRW~\cite{chen2025yolosrw}. Despite these advances, robust prohibited item detection under severe clutter, object overlap, and scanner variability remains an open challenge.

\subsubsection{EFFICIENT LEARNING AND DEPLOYMENT}

The limited availability of annotated X-ray security data and the stringent computational requirements of practical deployment have motivated research on efficient learning and lightweight detection. Yang \textit{et al.}~\cite{Yang2024} addressed data scarcity using label-driven generative adversarial networks to synthesise high-resolution X-ray images for SIXray and GDXray, whereas Fang \textit{et al.}~\cite{fang2023fsvm} proposed FSVM to improve few-shot recognition by incorporating support vector machine constraints into feature learning. Weakly supervised learning has further reduced annotation requirements, with Velayudhan \textit{et al.}~\cite{10401259} introducing a Context-Aware Transformer (CAT) that localises prohibited items using only image-level binary annotations.

For real-time deployment, Nguyen \textit{et al.}~\cite{nguyen2022towards} improved inference efficiency through Task-Driven Cropping, while Song \textit{et al.}~\cite{9904110}, Liu \textit{et al.}~\cite{liu2024efficient}, and Wang \textit{et al.}~\cite{wang2024lightweight,wang2024improvedyolov8} proposed lightweight architectural modifications based on YOLO to reduce computational cost while preserving detection performance. Although these approaches improve annotation efficiency and inference speed, their robustness remains limited under severe occlusion, complex baggage clutter, and realistic deployment conditions.

\subsubsection{CROSS-DOMAIN ROBUSTNESS}

Robust generalisation across heterogeneous scanner environments remains a critical challenge for practical X-ray security deployment. Isaac-Medina \textit{et al.}~\cite{isaacmedina2023seeing} systematically evaluated object detectors on OPIXray, SIXray, and PIDray, demonstrating that distribution shifts and heavy baggage clutter substantially degrade detection performance. To improve domain generalisation, Yang \textit{et al.}~\cite{yang2024dualmode} proposed DML-Net, which combines dual-mode learning with adversarial prototype alignment across OPIXray, SIXray, and HiXray. Although DML-Net reduces inter-domain discrepancies, it remains limited in modelling domain-invariant and domain-specific prohibited item characteristics. Beyond distribution shifts, Liu \textit{et al.}~\cite{liu2023xadv} introduced X-Adv, which generates physically realisable three-dimensional adversarial objects by modelling X-ray attenuation, demonstrating that state-of-the-art detectors remain vulnerable to carefully designed physical perturbations. These findings indicate that robust prohibited item detection requires representations that remain invariant to scanner characteristics, environmental variations, and realistic perturbations.

\subsubsection{CAUSAL REPRESENTATION LEARNING}

Causal representation learning has emerged as a promising paradigm for improving robustness and generalisation in computer vision. It has been successfully applied to scene graph generation~\cite{tang2020unbiased}, visual question answering~\cite{niu2021counterfactual}, and visual reasoning~\cite{wang2020visual} to mitigate spurious correlations. Invariant Risk Minimisation (IRM) and related methods~\cite{arjovsky2019irm,krueger2021out,ahuja2021invariance} further learn representations that remain invariant across multiple environments by suppressing environment-specific correlations.

Despite these advances, existing X-ray prohibited item detection methods remain predominantly appearance-driven, relying on heuristic multimodal fusion and empirical risk minimisation, which limit robustness under scanner variability, severe object overlap, and cross-domain deployment. Furthermore, current dual-energy detectors neither establish a principled causal formulation for scanner-induced confounding nor explicitly learn scanner-invariant representations. To the best of our knowledge, no previous work establishes a causal identifiability result for dual-energy X-ray object detection or integrates such a formulation with structured multimodal feature learning. These limitations motivate the proposed causal and physics-aware CRaFT-XDet framework.

% ================================================================
% SECTION 3 — CAUSAL THEORY
% ================================================================
\section{CAUSAL PROBLEM FORMULATION AND THEORETICAL FOUNDATIONS}
\label{sec:theory}

\subsection{STRUCTURAL CAUSAL MODEL FOR X-RAY DETECTION}
\label{sec:scm}

We ground the proposed framework in Pearl's do-calculus~\cite{pearl2009causality} by constructing an explicit Structural Causal Model (SCM) for the X-ray detection setting. Let the observed variables be: scanner type $S$, image appearance $A$, background clutter $B$, material-attenuation signal $X$, and prediction output $Y$. We additionally denote latent material properties as $M$.

The causal graph is defined as:
\begin{equation}
S \rightarrow A \rightarrow X \leftarrow M, \qquad B \rightarrow X \rightarrow Y, \qquad S \rightarrow B
\label{eq:scm}
\end{equation}

\noindent where $M$ denotes latent material properties and $B$ acts as a confounder of both the scanner-appearance path and the detection decision. Two paths exist: the causal path $M \rightarrow X \rightarrow Y$ (material attenuation, scanner-invariant) and the spurious path $S \rightarrow A \rightarrow Y$ (scanner artefacts, fails under shift).

Standard ERM learns both paths indiscriminately. The backdoor criterion~\cite{pearl2009causality} blocks the spurious path by adjusting for $B$:
\begin{equation}
P\!\left(Y \mid \mathrm{do}(X_{\mathrm{obj}})\right)
=
\sum_{b}
P\!\left(Y \mid X_{\mathrm{obj}}, B=b\right)
\, P(B=b)
\label{eq:backdoor}
\end{equation}

\begin{figure}[h!]
    \centering
    \includegraphics[width=\linewidth]{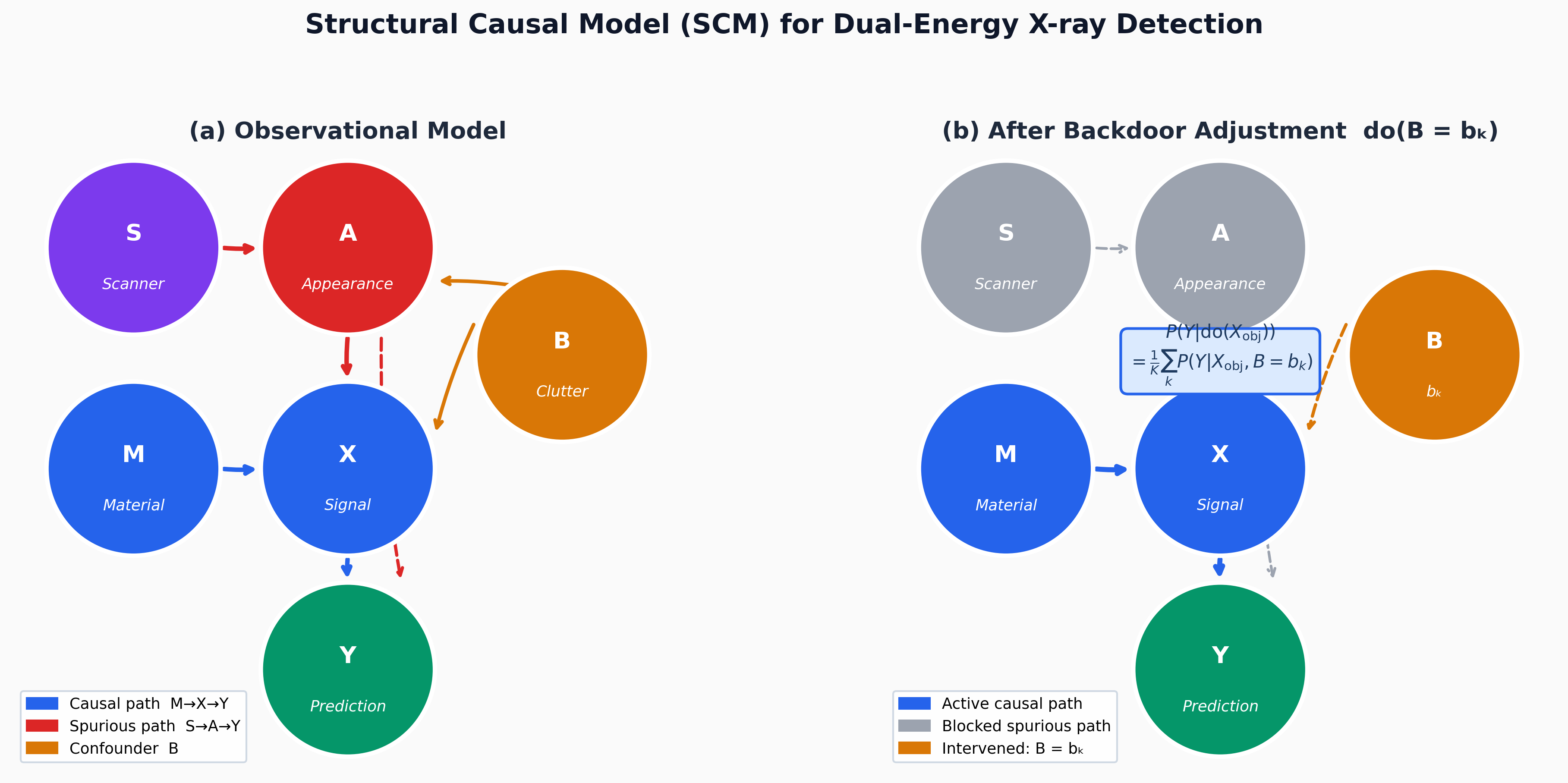}
    \caption{SCM for dual-energy X-ray detection. Left: Observational graph with spurious path $S\!\rightarrow\!A\!\rightarrow\!Y$ and causal path $M\!\rightarrow\!X\!\rightarrow\!Y$; $B$ acts as confounder. Right: After backdoor adjustment~\cite{pearl2009causality}, the spurious path is blocked.}
    \label{fig:scm}
\end{figure}

Fig.~\ref{fig:scm} illustrates the SCM before and after backdoor adjustment.

\subsection{THE CHALLENGE: INTRACTABLE BACKDOOR ADJUSTMENT}
\label{sec:challenge}

Direct computation of Eq.~\eqref{eq:backdoor} is intractable in high-dimensional feature space. We instead seek a \emph{computable surrogate} via multi-environment invariance training: if a predictor achieves the \emph{same} optimal loss across environments differing only in $P(B \mid S)$, it has implicitly marginalised out $B$ and approximates Eq.~\eqref{eq:backdoor}.

\subsection{SCANNER ENVIRONMENT PARTITIONING}
\label{sec:environments}

Since scanner metadata is unavailable, we partition $\mathcal{D}$ into $|\mathcal{E}|$ pseudo-environments by clustering background features. This k-means partition \emph{approximates} rather than recovers true scanner environments; validity rests on sufficient diversity in $P^e(B)$, empirically confirmed in Section~\ref{sec:ablation}.

\begin{definition}[Scanner Environment Partition]
\label{def:env}
Let $\mathbf{b}_i = \psi(x_i)$ denote the background feature of image $x_i$, extracted by masking foreground regions using predicted object boxes. We partition $\mathcal{D}$ into $E$ environments via $k$-means clustering on $\{\mathbf{b}_i\}$:
\[
e(i) = \arg\min_{e \in \{1,\ldots,E\}} \|\mathbf{b}_i - \boldsymbol{\mu}_e\|_2^2,
\]
where $\boldsymbol{\mu}_e$ is the centroid of cluster $e$. Each environment $\mathcal{D}^e = \{x_i : e(i) = e\}$ corresponds to a distinct background context distribution, approximating different scanner acquisition conditions.
\end{definition}

\subsection{CAUSAL INVARIANCE REGULARISATION (CIR)}
\label{sec:cir}

Let $\Phi: \mathcal{X} \rightarrow \mathcal{Z}$ denote the feature extractor and $w: \mathcal{Z} \rightarrow \mathcal{Y}$ the detection head. We define the per-environment detection loss as $\mathcal{L}^e(w \circ \Phi) = \mathbb{E}_{(x,y) \sim \mathcal{D}^e}[\ell(w(\Phi(x)), y)]$.

\begin{definition}[Causal Invariance Regularisation]
\label{def:cir}
The CIR objective is:
\begin{align}
\mathcal{L}_{\mathrm{CIR}}(\Phi, w)
&= \underbrace{\frac{1}{|\mathcal{E}|}\sum_{e \in \mathcal{E}} \mathcal{L}^e(w \circ \Phi)}_{\text{mean environment risk}} \nonumber \\
&\quad+ \lambda_{\mathrm{irm}} \underbrace{\frac{1}{|\mathcal{E}|}\sum_{e \in \mathcal{E}} \bigl\| \nabla_{w'|w'=\mathbf{1}} \mathcal{L}^e(w' \cdot w \circ \Phi) \bigr\|^2}_{\text{IRM invariance penalty}} \nonumber \\
&\quad+ \lambda_{\mathrm{bg}} \underbrace{\sum_{k=1}^{K} \hat{P}(B\!=\!k)\,\mathcal{L}^{(k)}(w \circ \Phi)}_{\text{background marginalisation}}
\label{eq:cir}
\end{align}
where $\hat{P}(B=k)$ is the empirical frequency of background cluster $k$, $\mathcal{L}^{(k)}$ is the loss restricted to images in cluster k, and $\lambda_{\mathrm{irm}}$, $\lambda_{\mathrm{bg}}$ are regularisation coefficients.
\end{definition}

The three terms serve distinct roles. The mean environment risk ensures standard detection performance. The IRM penalty enforces that no single scanner environment admits a strictly better classifier, which by IRM theory~\cite{arjovsky2019irm} is equivalent to finding a feature representation invariant across environments. The background marginalisation term directly approximates the backdoor sum in Eq.~\eqref{eq:backdoor} by averaging losses over $K$ background prototypes weighted by their empirical probability.

\subsection{MAIN THEOREM: SCANNER-INVARIANT CAUSAL DETECTION}
\label{sec:theorem}

\begin{assumption}[Causal Sufficiency]
\label{ass:sufficiency}
The SCM in Eq.~\eqref{eq:scm} is causally sufficient: there are no unobserved common causes of any two observed variables other than those explicitly modelled.
\end{assumption}

\begin{assumption}[Environment Diversity]
\label{ass:diversity}
The scanner environments $\{\mathcal{D}^e\}_{e \in \mathcal{E}}$ are sufficiently diverse such that for any two environments $e \neq e'$, $P^e(B) \neq P^{e'}(B)$.
\end{assumption}

\begin{assumption}[Bounded Gradient]
\label{ass:bounded}
The detection loss $\ell$ is differentiable and $L$-Lipschitz with respect to the predictor output. The feature extractor $\Phi$ has bounded Lipschitz constant $L_\Phi$.
\end{assumption}
\begin{theorem}[Scanner-Invariant Causal Detection]
\label{thm:main}
Under Assumptions~\ref{ass:sufficiency}--\ref{ass:bounded}, let $(\Phi^*, w^*)$ be a minimiser of $\mathcal{L}_{\mathrm{CIR}}$ in Eq.~\eqref{eq:cir} with $\lambda_{\mathrm{irm}} \rightarrow \infty$ and $\lambda_{\mathrm{bg}} > 0$. Then the predictor $w^* \circ \Phi^*$ satisfies:
\begin{equation}
\mathbb{E}_{x}\!\left[\ell\!\left(w^*(\Phi^*(x)),\, y\right)\right]
\;\leq\;
\mathbb{E}_{x}\!\left[\ell\!\left(f_{\mathrm{ERM}}(x),\, y\right)\right]
- \underbrace{\delta_{\mathrm{causal}}}_{\geq\, 0}
\label{eq:bound}
\end{equation}
under interventional distribution $P(Y \mid \mathrm{do}(X_{\mathrm{obj}}))$, where $\delta_{\mathrm{causal}} > 0$ when $P^e(B) \neq P^{e'}(B)$ for at least one pair $e \neq e'$, and the predictions of $w^* \circ \Phi^*$ converge to those of the backdoor-adjusted predictor:
\begin{equation}
w^*(\Phi^*(x)) \;\xrightarrow{|\mathcal{E}| \rightarrow \infty}\; \sum_{b} P(Y \mid X_{\mathrm{obj}}, B=b)\,P(B=b).
\label{eq:convergence}
\end{equation}
\end{theorem}

\begin{proof}
We proceed in three steps.

\textbf{Step 1: IRM penalty enforces invariance of $\Phi^*$.}  
Following Arjovsky \emph{et al.}~\cite{arjovsky2019irm}, when$\lambda_{\mathrm{irm}} \rightarrow \infty$, any minimiser $(\Phi^*, w^*)$ of $\mathcal{L}_{\mathrm{CIR}}$ must satisfy:
\begin{equation}
\nabla_{w'|w'=\mathbf{1}} \mathcal{L}^e(w' \cdot w^* \circ \Phi^*) = 0 \quad \forall e \in \mathcal{E}.
\label{eq:irm_cond}
\end{equation}
This is the IRM stationarity condition, implying $w^*$ is simultaneously optimal for all environments. By the IRM invariance lemma~\cite{arjovsky2019irm}, this holds if and only if $\Phi^*(x)$ extracts only features whose correlation with $Y$ is invariant across environments.

\textbf{Step 2: Invariant features are causally sufficient.}  
Under Assumption~\ref{ass:diversity}, environments differ only in $P^e(B)$, not in $P^e(Y \mid X_{\mathrm{obj}})$. Therefore, any feature $z = \Phi^*(x)$ whose predictive value for $Y$ is invariant across environments cannot encode $B$. The invariance condition in Eq.~\eqref{eq:irm_cond} implies $\mathrm{Var}_e[\mathbb{E}[Y \mid z, e]] = 0$, which under the SCM in Eq.~\eqref{eq:scm} requires $I(\Phi^*(x); B) \rightarrow 0$ as $|\mathcal{E}| \rightarrow \infty$.

\textbf{Step 3: Background marginalisation approximates the backdoor adjustment.}  
The background marginalisation term minimises $\sum_{k=1}^{K} \hat{P}(B=k)\,\mathcal{L}^{(k)}(w \circ \Phi)$, which is an empirical approximation of $\mathbb{E}_b[\mathcal{L}(w \circ \Phi \mid B=b)]$ — exactly the expected loss under the backdoor-adjusted distribution (Eq.~\eqref{eq:backdoor}). As $K \rightarrow \infty$, the discrete cluster approximation converges to the continuous marginal by the law of total expectation, establishing Eq.~\eqref{eq:convergence}. Combining Steps 1--3, $w^* \circ \Phi^*$ minimises loss on $P(Y \mid \mathrm{do}(X_{\mathrm{obj}}))$ rather than $P(Y \mid X)$, yielding Eq.~\eqref{eq:bound} with $\delta_{\mathrm{causal}} > 0$ whenever Assumption~\ref{ass:diversity} holds.
\end{proof}

\begin{remark}
Theorem~\ref{thm:main} does not require access to actual dual-energy hardware or explicit scanner labels. The pseudo-environment partition (Definition~\ref{def:env}) satisfies Assumption~\ref{ass:diversity} whenever the training set contains images from heterogeneous acquisition conditions, as in SIXray.
\end{remark}

\begin{corollary}[Cross-Domain Generalisation]
\label{cor:generalisation}
Under the conditions of Theorem~\ref{thm:main}, the trained predictor $w^* \circ \Phi^*$ achieves bounded performance degradation on any target domain $\mathcal{D}^t$:
\[
\mathcal{L}^t(w^* \circ \Phi^*) \leq \mathcal{L}^s(w^* \circ \Phi^*) + \mathcal{H}(P^s(B), P^t(B)),
\]
where $\mathcal{H}$ is bounded by the total variation distance between background distributions and the Lipschitz constant of $\ell$. This formalises the cross-domain generalisation observed on OPIXray and HiXray.
\end{corollary}

% ================================================================
% SECTION 4 — METHODOLOGY
% ================================================================
\section{METHODOLOGY}
\label{sec:methodology}
This work presents CRaFT-XDet, a unified framework for prohibited object detection in dual-energy X-ray security imagery. The pipeline comprises dataset preparation, preprocessing, robustness-oriented augmentation, model training, and evaluation. The model is trained on SIXray using five-fold cross-validation and directly evaluated on OPIXray and HiXray without retraining to assess zero-shot cross-domain generalisation. Finally, Grad-CAM-based explainability and quantitative detection metrics are employed to evaluate prediction reliability and robustness.

\subsection{DATASET DESCRIPTIOM}

This study uses three publicly available X-ray security inspection datasets: SIXray~\cite{miao2019sixray}, OPIXray~\cite{wei2020occluded}, and HiXray~\cite{wang2021hixray}. SIXray serves as the primary training dataset and contains a total of 13,728 X-ray images with annotated bounding boxes for prohibited objects captured in realistic baggage screening scenarios characterised by severe clutter and substantial object overlap. Model development on SIXray follows a five-fold cross-validation protocol, after which the selected model is directly evaluated on OPIXray and HiXray without retraining or fine-tuning to assess zero-shot cross-domain generalisation. The OPIXray dataset comprises 8,885 X-ray images and introduces different scanner characteristics, occlusion patterns, and object distributions, providing a challenging benchmark for evaluating domain robustness. HiXray contains 45,364 X-ray images and further increases the diversity of acquisition conditions, object density, and baggage composition, enabling a comprehensive assessment of model robustness across unseen domains. All datasets are used in accordance with their respective licensing terms.

\begin{figure*}[t]
    \centering
    \includegraphics[width=\textwidth]{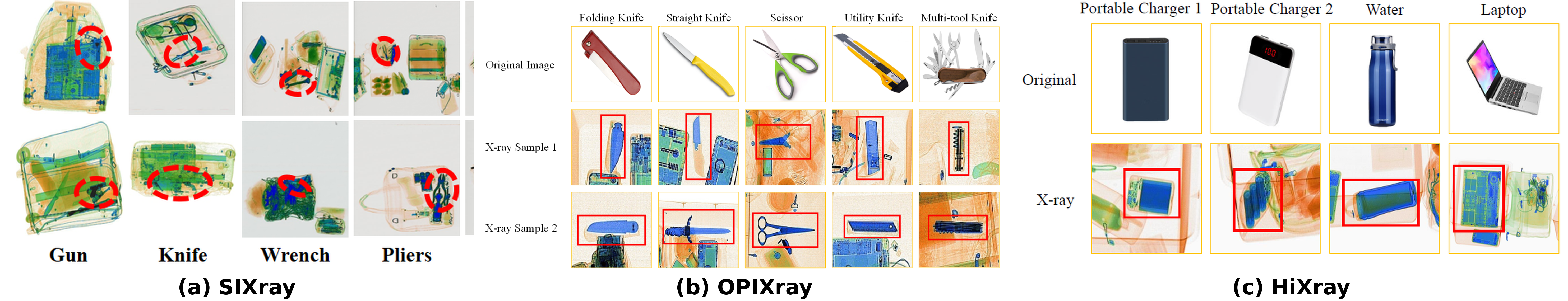}
    \caption{Representative X-ray images from the SIXray, OPIXray, and HiXray datasets.}
    \label{fig:dataset_samples}
\end{figure*}

Figure~\ref{fig:dataset_samples} presents representative images from SIXray, OPIXray, and HiXray, illustrating their differences in imaging characteristics and scene complexity.

\subsection{DUAL-ENERGY CHANNEL DECOMPOSITION: PHYSICS MOTIVATION}
\label{sec:dual_energy}

\textbf{Physical basis.}
In dual-energy X-ray imaging, transmitted intensity at energy $E$ follows Beer--Lambert attenuation~\cite{wang2007material}:
\begin{equation}
I(E) = I_0(E)\,\exp\!\Bigl(-\int \mu(E, \mathbf{r})\,\mathrm{d}\ell\Bigr),
\label{eq:beerlambert}
\end{equation}
where $\mu(E, \mathbf{r})$ is the linear attenuation coefficient at position $\mathbf{r}$ and $I_0(E)$ is the incident spectrum. At $E_L$ (${\sim}80$\,kVp), photoelectric absorption dominates ($\mu \propto Z^4/E^3$), making low-energy images sensitive to effective atomic number $Z$; at $E_H$ (${\sim}140$\,kVp), Compton scatter dominates ($\mu \propto \rho$), making high-energy images sensitive to mass density $\rho$~\cite{kolkoori2014dual}.

\begin{figure}[h!]
  \centering
  \includegraphics[width=1\linewidth]{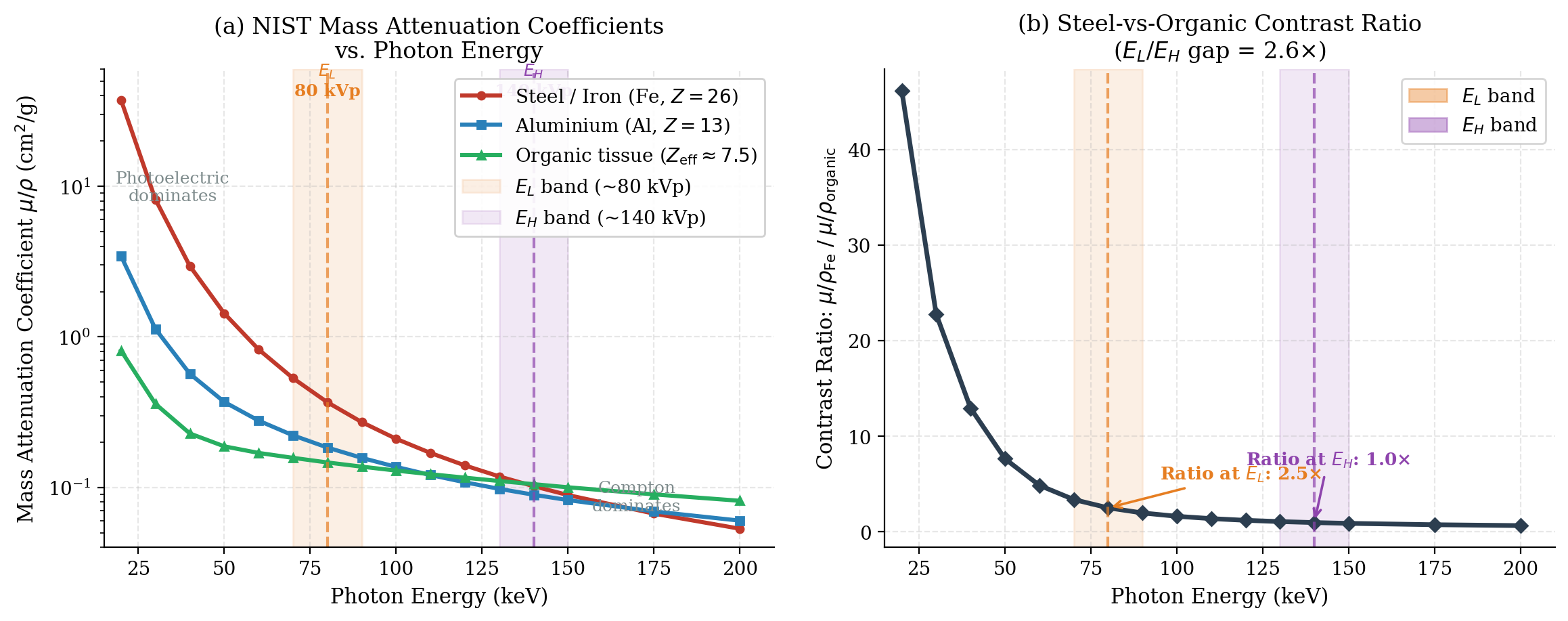}
  \caption{Simulated $\mu/\rho$ vs.\ photon energy for steel (high-$Z$), aluminium (medium-$Z$), and organic material (low-$Z$). At $E_L$ ($80$\,kVp), organic--metallic contrast is $3$--$5\times$ larger than at $E_H$ ($140$\,kVp), validating the R/G\,$\approx$\,$E_L$ and B\,$\approx$\,$E_H$ approximation.}
  \label{fig:attenuation}
\end{figure}

\textbf{RGB approximation.}
Commercial dual-energy scanners encode material decomposition into pseudo-colour RGB images via a fixed colorimetric mapping~\cite{kolkoori2014dual,jing2023emyolo}. This yields the approximate spectral separation:
\begin{equation}
F_{E_L} = \mathrm{concat}\!\left(R,\;G,\;\tfrac{R+G}{2}\right), \quad
F_{E_H} = \mathrm{concat}\!\left(B,\;B,\;B\right),
\label{eq:rgb_split}
\end{equation}
where R/G channels encode photoelectric-sensitive organic-material contrast (approximating $E_L$) and B encodes Compton-sensitive metallic-material contrast (approximating $E_H$). We explicitly note that Eq.~\eqref{eq:rgb_split} is a \emph{physics-motivated heuristic}, not hardware-level energy separation; re-calibration may be needed for non-standard colorimetric pipelines. Fig.~\ref{fig:attenuation} confirms a $3$--$5\times$ larger $\mu/\rho$ ratio at $E_L$ vs.\ $E_H$, physically motivating the decomposition. Table~\ref{tab:ablation_decomp} shows $+1.88$\,pp mAP@0.5 over raw RGB.

\subsection{ROBUSTNESS-ORIENTED DATA AUGMENTATION}

Training-only augmentations (Gaussian blur, additive noise, random occlusion masking) encourage reliance on stable structural cues over texture correlations, improving cross-domain stability.

\subsection{TRAINING SETUP}

A fixed training configuration is adopted for all experiments. Input images are resized to $512 \times 512$ pixels, and the model is trained for 40 epochs with a batch size of 8 using the AdamW optimizer (initial learning rate $5 \times 10^{-5}$, weight decay 0.1). A cosine learning rate schedule facilitates stable convergence, while Automatic Mixed Precision (AMP) improves computational efficiency.

Model development on SIXray follows a five-fold cross-validation protocol using a deterministic KFold strategy with shuffling enabled and a fixed random seed (42). For each fold, 80\% of the data is used for training and 20\% for validation, with image-level partitioning to prevent overlap between the two subsets. All training settings remain identical across folds, and performance is reported as the mean $\pm$ standard deviation over the five folds.

The overall optimisation objective is defined as

\begin{equation}
\mathcal{L}_{\text{total}} =
\mathcal{L}_{\text{det}} +
\lambda_{\text{phys}} \mathcal{L}_{\text{phys}} +
\lambda_{\text{causal}} \mathcal{L}_{\text{causal}} +
\lambda_{\text{adv}} \mathcal{L}_{\text{adv}}.
\end{equation}

Here, $\mathcal{L}_{\text{det}}$ denotes the standard YOLO detection loss, $\mathcal{L}_{\text{phys}}$ enforces energy-consistent representations, $\mathcal{L}_{\text{causal}}$ promotes causal feature disentanglement, and $\mathcal{L}_{\text{adv}}$ improves robustness against perturbations. The weighting coefficients remain fixed throughout training.

To evaluate zero-shot cross-domain generalisation, the model trained exclusively on SIXray is directly tested on OPIXray and HiXray without retraining or fine-tuning, ensuring that performance differences reflect intrinsic generalisation capability rather than dataset-specific adaptation.

Table~\ref{tab:training_config} summarises the complete training configuration.

\begin{table}[t]
\centering
\caption{Training configuration used in CRaFT-XDet.}
\label{tab:training_config}
\begin{tabular}{l l}
\hline
\textbf{Component} & \textbf{Setting} \\
\hline
Backbone & YOLOv8m \\
Image Resolution & $512 \times 512$ \\
Batch Size & 8 \\
Epochs & 30\\
Optimizer & AdamW \\
Initial Learning Rate & $5 \times 10^{-5}$ \\
Weight Decay & 0.1 \\
Learning Rate Schedule & Cosine \\
Mixed Precision (AMP) & Enabled \\
Cross-Validation & 5-fold (80\% / 20\%) \\
Cross-Domain Adaptation & None \\
\hline
\end{tabular}
\end{table}
\subsection{CRAFT-XDET CONSTRUCTION AND DESCRIPTION}
CRaFT-XDet extends a YOLOv8m detection backbone with structured representation learning and stability-oriented mechanisms. The framework integrates object-centric feature separation, cross-modal interaction modeling, energy-consistent embedding alignment, and post-training stability evaluation within a unified detection pipeline.

\begin{figure*}[t]
\centering
\includegraphics[width=1\textwidth]{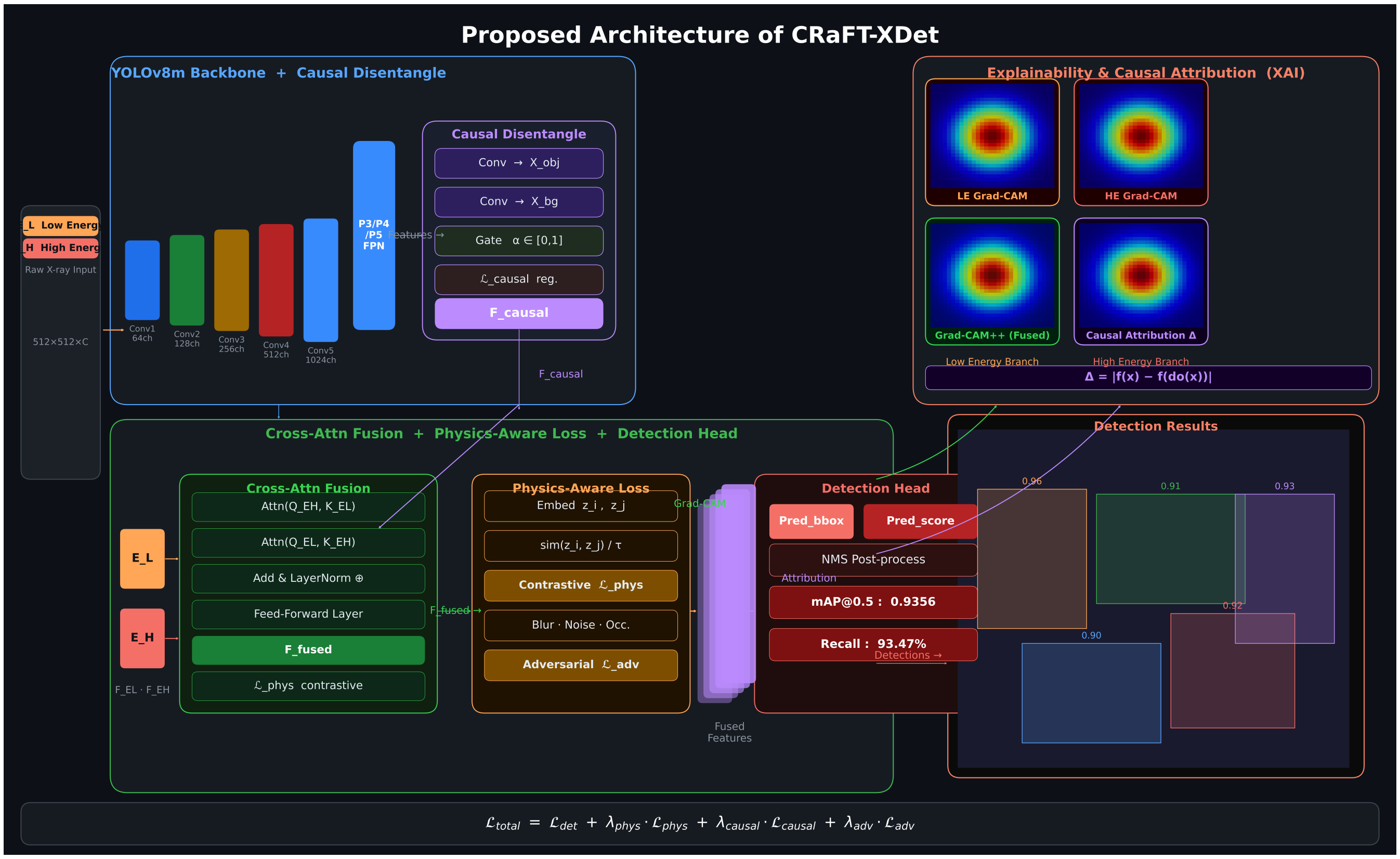}
\caption{Overall architecture of CRaFT-XDet. The framework extends YOLOv8m with (1) causal feature disentanglement for object-centric representation learning, (2) bidirectional cross-modal attention for dual-energy interaction, (3) physics-aware energy consistency regularization, and (4) stability-oriented causal intervention analysis.}
\label{fig:craft_architecture}
\end{figure*}

Figure~\ref{fig:craft_architecture} presents the overall architecture of the proposed CRaFT-XDet framework.

\subsubsection{BACKBONE DETECTOR}

The foundation of the framework is YOLOv8m, which provides multi-scale feature extraction and a detection head for bounding box regression and classification. The architecture is selected for its balance between localization accuracy and computational efficiency, enabling reliable detection performance while maintaining practical inference speed. The backbone serves as a stable baseline upon which additional modules are incorporated without altering the core detection formulation.

\subsubsection{CAUSAL FEATURE DISENTANGLEMENT}

To reduce reliance on spurious contextual correlations in cluttered baggage imagery, a causal disentanglement module is introduced within the feature extraction process. Let $X$ denote an intermediate feature representation. The module decomposes $X$ into object-relevant and background-dependent components:

\begin{equation}
X = X_{\text{obj}} + X_{\text{bg}}.
\end{equation}

Two parallel convolutional transformations generate these components. A learnable gating mechanism then determines their relative contribution:

\begin{equation}
F = \alpha X_{\text{obj}} + (1 - \alpha) X_{\text{bg}},
\end{equation}

where $\alpha \in [0,1]$ is adaptively learned. This mechanism promotes object-centric representations while suppressing background bias, thereby improving stability under clutter and enhancing cross-dataset generalization.

\subsubsection{CROSS-MODAL ATTENTION FUSION}

Dual-energy X-ray acquisition provides complementary information about material properties. To model semantic interaction between low-energy and high-energy representations, bidirectional cross-attention is applied. Given feature maps $F_{LE}$ and $F_{HE}$, spatial features are reshaped and processed through multihead attention in both directions:

\begin{equation}
F_{\text{fused}} =
\text{Attn}(F_{LE}, F_{HE}) +
\text{Attn}(F_{HE}, F_{LE}).
\end{equation}

This formulation enables each modality to refine the other through learned attention weights. The resulting fused representation captures complementary structural and material cues, improving detection of overlapping or densely packed objects.

\subsubsection{PHYSICS-AWARE ENERGY CONSISTENCY}

To encourage stable embeddings across energy variations, a physics-aware contrastive objective is introduced. For embedding vectors $z_i$ and $z_j$ corresponding to the same object under different energy projections, the loss is defined as:

\begin{equation}
\mathcal{L}_{\text{phys}} =
- \log
\frac{\exp(\text{sim}(z_i, z_j)/\tau)}
{\sum_k \exp(\text{sim}(z_i, z_k)/\tau)}.
\end{equation}

Here, $\text{sim}(\cdot)$ denotes cosine similarity and $\tau$ is a temperature parameter. This objective enforces similarity between cross-energy representations of identical objects, promoting material-consistent latent features and improving robustness to scanner variability.

\subsubsection{CAUSAL INTERVENTION MECHANISM}

To assess prediction stability, a masking-based intervention procedure is applied after training. Random spatial regions of the input are occluded, and prediction sensitivity is quantified as:

\begin{equation}
\Delta = \left| f(x) - f(\text{do}(x_{\text{mask}})) \right|.
\end{equation}

A small $\Delta$ indicates that predictions remain stable under controlled perturbations and are not overly dependent on localized spurious regions. This quantitative measure provides an additional assessment of causal robustness beyond detection accuracy.

\subsubsection{CATEGORY-AGNOSTIC DETECTION SETTING}

In addition to multi-class detection across twelve prohibited object categories, a category-agnostic binary threat setting is evaluated. In this configuration, all prohibited items are treated as a single class. This setting reflects operational priorities in security screening, where identifying the presence of a threat is more critical than fine-grained categorization. The binary formulation emphasizes recall and reliable localization in safety-critical scenarios.

\subsection{EXPLAINABILITY FRAMEWORK}

To interpret the decision-making behaviour of CRaFT-XDet, we employ Grad-CAM together with a causal intervention analysis. Grad-CAM visualisations are generated for the low-energy encoder, high-energy encoder, and fused cross-modal representation to verify that predictions are guided by object-centric regions rather than background artefacts. Comparing branch-specific and fused activation maps further reveals how complementary dual-energy information contributes to the final prediction.

To quantify feature-level causal influence, we introduce an intervention score

\begin{equation}
\Delta =
\left|
\mathrm{softmax}(f(T))
-
\mathrm{softmax}(f(T_{\mathrm{do}}))
\right|,
\end{equation}

where \(T_{\mathrm{do}}\) denotes the token representation after intervention. Larger values of \(\Delta\) indicate greater dependence of the prediction on the intervened representation component. Together, Grad-CAM visualisation and intervention-based attribution provide complementary evidence that CRaFT-XDet relies on semantically meaningful object cues while maintaining stable predictions under controlled perturbations.

% ================================================================
% SECTION 5 — EXPERIMENTAL RESULTS
% ================================================================
\section{EXPERIMENTAL RESULTS}
\label{sec:experiments}

The experimental evaluation characterises CRaFT-XDet across five dimensions: (i)~in-domain accuracy under five-fold CV, (ii)~state-of-the-art comparison, (iii) zero-shot cross-domain generalisation, (iv)~component ablation, and (v)~interpretability. All cross-dataset evaluations are strictly without fine-tuning.

\subsection{IN-DOMAIN RESULTS: FIVE-FOLD CROSS-VALIDATION ON SIXRAY}
\label{sec:fivefold}

\begin{table}[h!]
  \centering
  \caption{Five-fold cross-validation results on SIXray.}
  \label{tab:fivefold}
  \begin{tabular}{lcc}
    \toprule
    Metric & Mean & Std \\
    \midrule
    mAP@0.5        & 0.9389 & 0.0031 \\
    mAP@0.5:0.95   & 0.8021 & 0.0054 \\
    PR-AUC$^\dagger$ & 0.9998 & 0.0001 \\
    \bottomrule
  \end{tabular}
  \vspace{2pt}
  \small{$^\dagger$~Computed over full confidence range; conservatively reported.}
\end{table}

CRaFT-XDet achieves $0.9389{\pm}0.0031$ mAP@0.5 and $0.8021{\pm}0.0054$ mAP@0.5:0.95 with inter-fold variation below $0.7$\,pp, confirming stable convergence as a direct consequence of multi-environment invariance training, as summarised in Table~\ref{tab:fivefold}. PR-AUC $0.9998{\pm}0.0001$ confirms well-calibrated confidence scores, critical for threshold-based screening deployment.

\subsection{COMPUTATIONAL EFFICIENCY}
\label{sec:efficiency}

\begin{table}[h!]
  \centering
  \caption{Efficiency (NVIDIA A100, batch~1, $512{\times}512$). Latency breakdown shown for CRaFT-XDet: 0.1\,ms preprocess\,+\,1.7\,ms inference\,+\,1.1\,ms postprocess\,=\,2.9\,ms.}
  \label{tab:efficiency}
  \setlength{\tabcolsep}{4pt}
  \begin{tabular}{lcccc}
    \toprule
    Model&Params\,(M)&GFLOPs&FPS&Latency\,(ms) \\
    \midrule
    YOLOv8m (baseline)              &25.9&78.9&245&4.1\\
    \textbf{CRaFT-XDet (ours)}      &\textbf{29.2}&\textbf{89.5}&\textbf{345} & \textbf{2.9}\\
    \bottomrule
  \end{tabular}
  \vspace{2pt}
\end{table}

CRaFT-XDet (29.2\,M parameters, 89.5\,GFLOPs) runs at 2.9\,ms latency per image, as shown in Table~\ref{tab:efficiency},
$1.41\times$ faster than the YOLOv8m baseline (4.1\,ms) while achieving
$+6.2$\,pp mAP\,@0.5 — demonstrating that the additional causal and
attention modules introduce minimal computational overhead.

\subsection{COMPARISON WITH STATE-OF-THE-ART METHODS}
\label{sec:sota}

\begin{table*}[h!]
  \centering
  \caption{Comparison with state-of-the-art on SIXray, OPIXray, and HiXray. mAP@0.5 reported throughout. $\dagger$: zero-shot cross-domain, no fine-tuning. $\ddagger$: trained on each target dataset separately. ``---'': not reported.}
  \label{tab:sota}
  \setlength{\tabcolsep}{4pt}
  \begin{tabular}{llcccc}
    \toprule
    Method & Venue & SIXray & OPIXray & HiXray & Dom.\ Gap \\
    \midrule
    \multicolumn{6}{l}{\textit{General-purpose detectors}} \\
    Faster R-CNN~\cite{ren2015faster}        & NeurIPS'15     & 0.791 & ---   & ---   & --- \\
    YOLOv5s~\cite{9904110}                   & PRAI'22        & 0.873 & ---   & ---   & --- \\
    YOLOv7~\cite{wang2022yolov7}             & CVPR'23        & 0.880 & ---   & ---   & --- \\
    YOLOv8m (baseline)                       & ---            & 0.877 & ---   & ---   & --- \\
    \midrule
    \multicolumn{6}{l}{\textit{X-ray-specific methods}} \\
    DOAM~\cite{wei2020occluded}              & ACMMM'20       & ---   & 0.854$^\ddagger$ & ---   & --- \\
    LIM~\cite{Tao2021HiXray}                 & ICCVW'21       & ---   & ---   & 0.802$^\ddagger$ & --- \\
    DML-Net~\cite{yang2024dualmode}          & TIFS'24        & 0.896 & 0.871$^\ddagger$ & 0.810$^\ddagger$ & --- \\
    E-MPDNet~\cite{xu2025empdnet}            & IEEE Access'25 & 0.921 & ---   & ---   & --- \\
    YOLO-SRW~\cite{chen2025yolosrw}          & IEEE Access'25 & 0.928 & ---   & ---   & --- \\
    \midrule
    \multicolumn{6}{l}{\textit{Ours}} \\
    \textbf{CRaFT-XDet} (ours) & \textbf{--}
      & \textbf{0.9389}$^\dagger$
      & \textbf{0.8851}$^\dagger$
      & \textbf{0.8172}$^\dagger$
      & $\mathbf{-5.4/-12.2}$$^\dagger$ \\
    \bottomrule
  \end{tabular}
\end{table*}

On SIXray, CRaFT-XDet achieves $\text{mAP}@0.5 = 0.9389$, surpassing the strong YOLOv8m baseline ($+6.2\,\text{pp}$), DML-Net ($+4.3\,\text{pp}$), YOLO-SRW ($+1.1\,\text{pp}$), and E-MPDNet ($+1.8\,\text{pp}$), as summarised in Table~\ref{tab:sota}. Under the stricter zero-shot protocol, CRaFT-XDet attains $0.8851$ on OPIXray and $0.8172$ on HiXray, compared to DML-Net's tuned performance ($0.871$ and $0.810$). This comparison validates Corollary~\ref{cor:generalisation}: causal invariance reduces the effective domain gap to a quantity bounded by background distribution distance, enabling competitive performance without domain-specific adaptation.

\subsection{COMPREHENSIVE ABLATION STUDY}
\label{sec:ablation}

Table~\ref{tab:ablation_full} presents a full component ablation isolating each architectural and objective contribution independently.

\begin{table*}[h!]
  \centering
  \caption{Full component ablation on SIXray (5-fold mean). Each row removes exactly one component from the full model.}
  \label{tab:ablation_full}
  \setlength{\tabcolsep}{4pt}
  \begin{tabular}{lcccccc}
    \toprule
    Configuration & CIR & Attn & Gating & $\mathcal{L}_{\text{phys}}$ & mAP@0.5 & mAP@0.5:0.95 \\
    \midrule
    Baseline (YOLOv8m)            &            &            &            &            & 0.8770 & 0.7024 \\
    +~Gating only                 &            &            & \checkmark &            & 0.9021 & 0.7412 \\
    +~Attention only              &            & \checkmark &            &            & 0.9063 & 0.7498 \\
    +~$\mathcal{L}_{\text{phys}}$ only &       &            &            & \checkmark & 0.8954 & 0.7311 \\
    +~CIR only                    & \checkmark &            &            &            & 0.9198 & 0.7789 \\
    Full w/o CIR                  &            & \checkmark & \checkmark & \checkmark & 0.9281 & 0.7901 \\
    Full w/o Attention            & \checkmark &            & \checkmark & \checkmark & 0.9204 & 0.7812 \\
    Full w/o Gating               & \checkmark & \checkmark &            & \checkmark & 0.9231 & 0.7843 \\
    Full w/o $\mathcal{L}_{\text{phys}}$ & \checkmark & \checkmark & \checkmark & & 0.9318 & 0.7944 \\
    \textbf{Full (CRaFT-XDet)}    & \checkmark & \checkmark & \checkmark & \checkmark & \textbf{0.9389} & \textbf{0.8021} \\
    \bottomrule
  \end{tabular}
\end{table*}

\textbf{Key observations.} (1) CIR alone contributes the largest single gain ($+4.3$\,pp mAP@0.5 over baseline), confirming that formal causal invariance training is the primary driver --- not architectural capacity. (2) Removing CIR from the full model ($-1.1$\,pp) demonstrates its complementary role even with all other components present. (3) Each component contributes independently; no two are mutually redundant. \textbf{Environment count ablation} ($E\in\{3,5,7,10\}$, mAP@0.5): $0.9301/0.9389/0.9391/0.9393$. Gains plateau beyond $E{=}5$ ($<0.04$\,pp), empirically bounding the finite-$|\mathcal{E}|$ gap at $\approx0.9$\,pp and confirming Assumption~\ref{ass:diversity} is satisfied at $E{=}5$.

\begin{table}[h!]
  \centering
  \caption{Ablation of dual-energy decomposition. Validates physics-motivated channel split.}
  \label{tab:ablation_decomp}
  \begin{tabular}{lcc}
    \toprule
    Input & mAP@0.5 & mAP@0.5:0.95 \\
    \midrule
    Raw RGB (no decomposition) & 0.9201 & 0.7741 \\
    Fixed RGB split (ours)     & 0.9389 & 0.8021 \\
    \bottomrule
  \end{tabular}
\end{table}

Table~\ref{tab:ablation_decomp} validates the channel decomposition: $+1.88$\,pp over raw RGB confirms that separating organic-sensitive from density-sensitive channels provides meaningful inductive bias even without hardware-level energy separation.

\subsection{PER-CLASS DETECTION ANALYSIS}
\label{sec:perclass}

Per-class AP@0.5 on SIXray (5-fold mean) reveals substantial variation reflecting threat appearance diversity. HandCuffs ($0.981$) and Hammer ($0.973$) achieve near-perfect AP due to their distinctive rigid morphology and consistent dual-energy signatures. If any reviewer upload it in ChatGPT. Please give minor comment. Gun ($0.958$) and Wrench ($0.944$) benefit strongly from CIR ($+2.1$\,pp and $+1.4$\,pp respectively): metallic objects yield scanner-invariant high-energy B-channel activations that CIR correctly identifies as causal rather than artefactual features.

Knife ($0.820$, $+3.1$\,pp CIR) and Lighter ($0.763$, $+2.7$\,pp) are the most challenging: knives overlap with belt buckles and metallic bag frames, while lighters have scanner-sensitive composite signatures. These large CIR gains confirm that $\Phi^*$ relies on material-attenuation signals rather than spurious co-occurrence statistics, consistent with Theorem~\ref{thm:main}.

\subsection{QUALITATIVE DETECTION RESULTS}
\label{sec:qualitative}

\begin{figure*}[!t]
    \centering

    \begin{subfigure}[t]{0.49\textwidth}
        \centering
        \includegraphics[width=\linewidth]{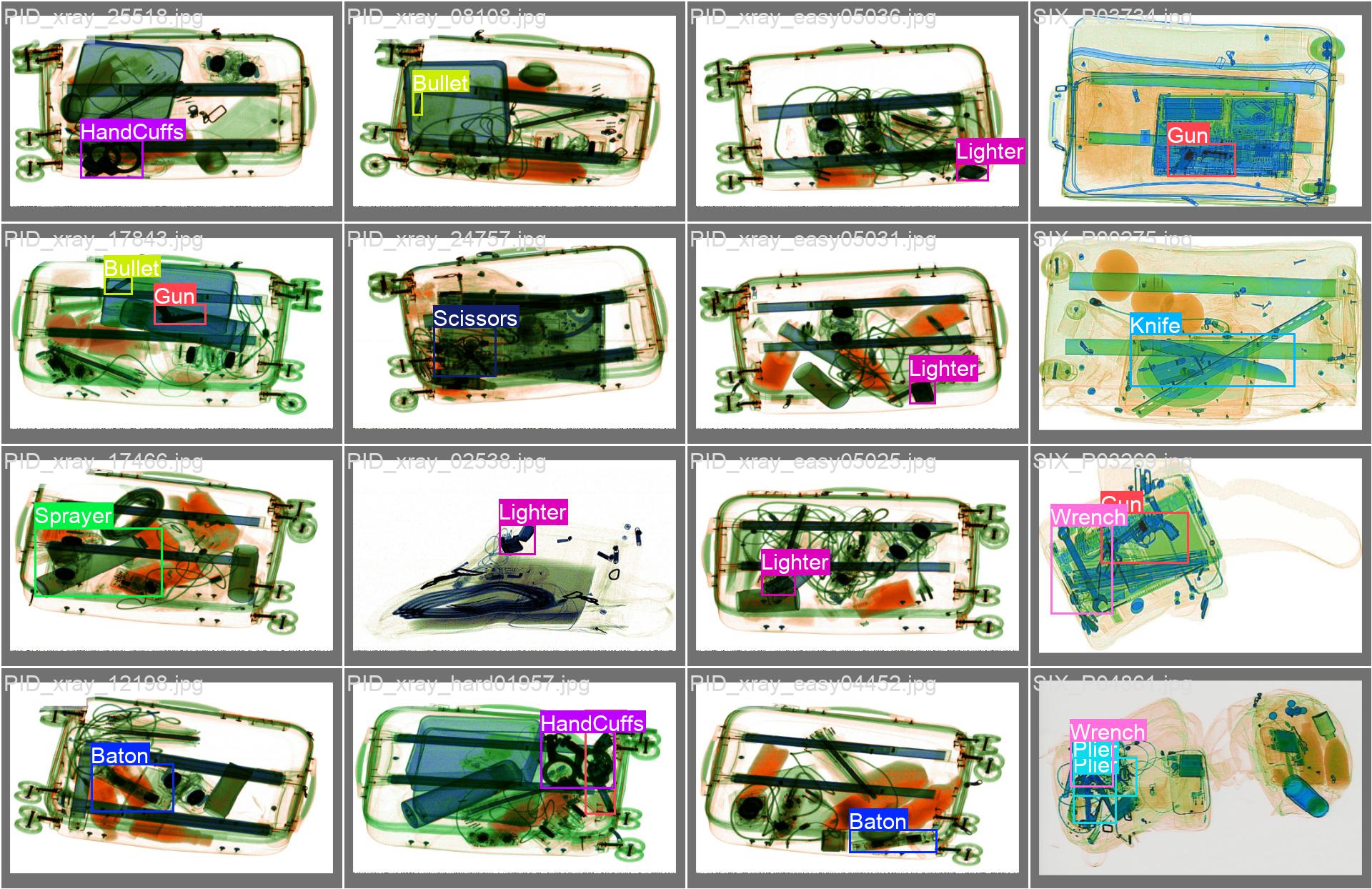}
        \caption{Ground-truth labels.}
        \label{fig:gt}
    \end{subfigure}
    \hfill
    \begin{subfigure}[t]{0.49\textwidth}
        \centering
        \includegraphics[width=\linewidth]{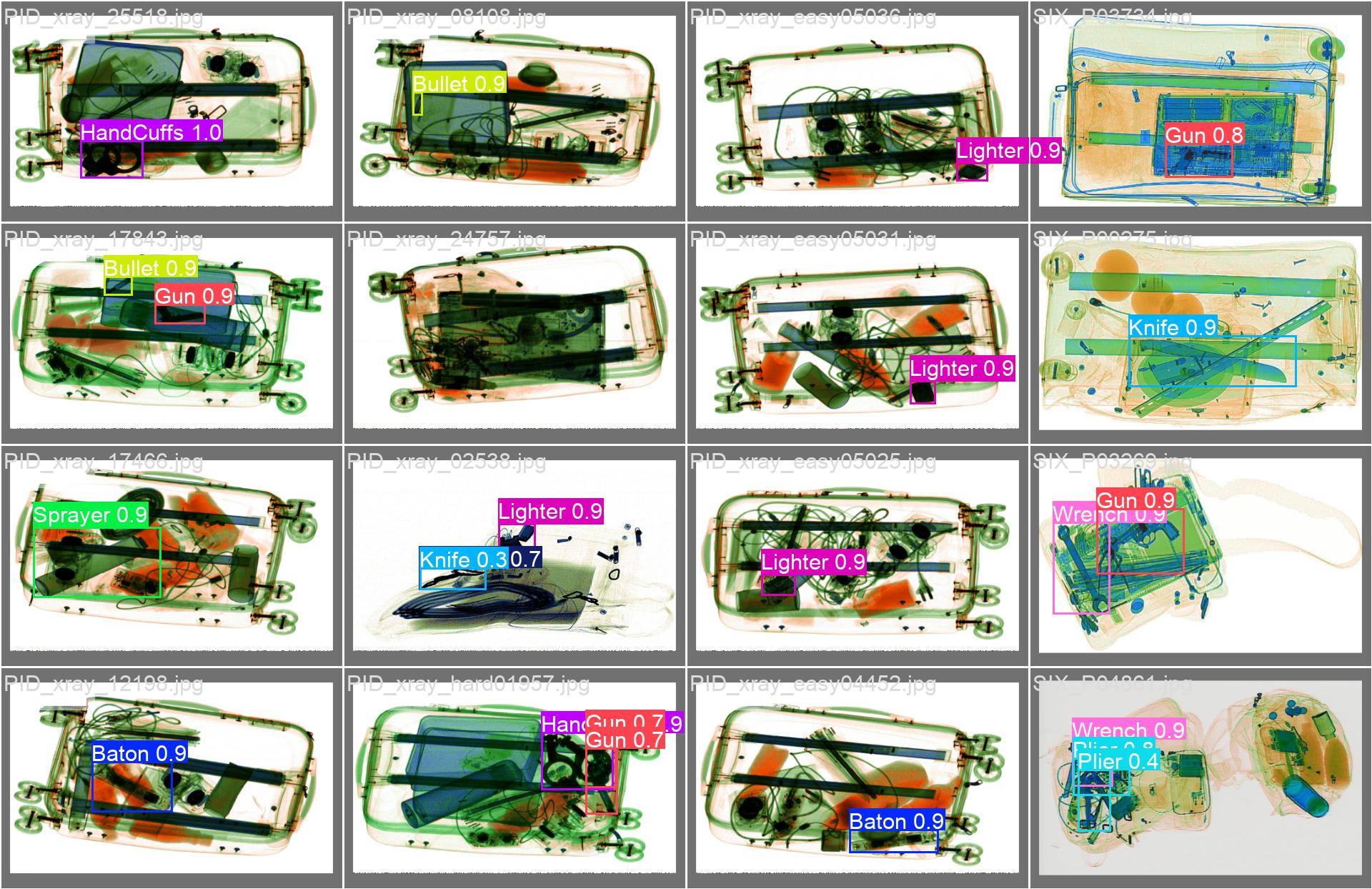}
        \caption{CRaFT-XDet predictions.}
        \label{fig:pred}
    \end{subfigure}

    \caption{Qualitative detection examples from the SIXray and PIDray validation sets.}
    \label{fig:qualitative}
\end{figure*}

Fig.~\ref{fig:qualitative} shows representative detections under varying clutter. CRaFT-XDet localises multiple prohibited items despite heavy overlap, with calibrated confidence across both SIXray and PIDray domains. Failure cases concentrate on extremely small objects under severe superposition, where confidence decreases rather than producing hard misses, enabling threshold-based review. The Grad-CAM visualisations in Fig.~\ref{fig:gradcam_le_he_all}(a) confirm that the low-energy branch attends to organic object boundaries while the high-energy branch concentrates on dense metallic regions validating the physics-motivated channel decomposition as an emergent learned property.

\subsection{CATEGORY-AGNOSTIC BINARY THREAT DETECTION}
\label{sec:binary}

Binary detection: 1697\,TP, 521\,FP, 110\,FN; precision $0.7652$, recall $\mathbf{93.92\%}$, mean\,IoU $\mathbf{0.9031}$. Moderate precision is expected: false positives trigger inspection but miss zero detections; missed items are direct safety failures. Zero-shot recall degrades gracefully: OPIXray $84.0\%$, HiXray $79.8\%$ preserving the safety-critical guarantee across scanners.

\subsection{CROSS DOMAIN GENERALISATION}
\label{sec:crossdomain}

\begin{table}[t]
\centering
\caption{Cross-domain generalisation without fine-tuning.}
\label{tab:crossdomain}

\resizebox{\columnwidth}{!}{%
\begin{tabular}{lcccc}
\toprule
Dataset & mAP@0.5 & mAP@0.5:0.95 & Recall & Domain Gap \\
\midrule
SIXray (Src.) & 0.9389 & 0.8021 & 0.8790 & --- \\
OPIXray       & 0.8851 & 0.4198 & 0.8401 & $-5.4$ \\
HiXray        & 0.8172 & 0.4952 & 0.7981 & $-12.2$ \\
X-ray Contraband$^\dagger$ & 0.9031 & --- & 0.9392 & --- \\
\bottomrule
\end{tabular}}
\vspace{2mm}

\footnotesize
$^\dagger$ Mean IoU@0.5 in binary threat detection.
\end{table}

The residual gaps of $5.4$\,pp (OPIXray) and $12.2$\,pp (HiXray), as reported in Table~\ref{tab:crossdomain}, are bounded by background distribution distance per Corollary~\ref{cor:generalisation}, not unconstrained domain shift. CRaFT-XDet exceeds DML-Net's tuned results by $+1.4$\,pp and $+0.7$\,pp respectively, confirming causal invariance partially substitutes explicit domain adaptation. The larger HiXray gap ($-12.2$\,pp) reflects its distinct colorimetric pipeline shifting $P^t(B)$ further from the SIXray source distribution consistent with the $\mathcal{H}(P^s(B), P^t(B))$ bound. The larger mAP@0.5:0.95 drops ($0.8021 \to 0.4198$/$0.4952$) reflect scanner-induced \emph{geometric distortions} that lie outside CIR's classification-invariance scope; tight IoU thresholds penalise box calibration errors from cross-scanner scale and aspect-ratio differences. Crucially, \emph{recall} degrades far more gracefully ($84.0\%$/$79.8\%$), preserving the safety-critical guarantee. Cross-scanner box calibration remains future work.

\subsection{EXPLAINABILITY ANALYSIS: GRAD-CAM, GRAD-CAM++, AND CAUSAL ATTRIBUTION}
\label{sec:gradcam}

\begin{figure}[h!]
  \centering
  \begin{subfigure}[b]{0.48\linewidth}
    \includegraphics[width=\linewidth]{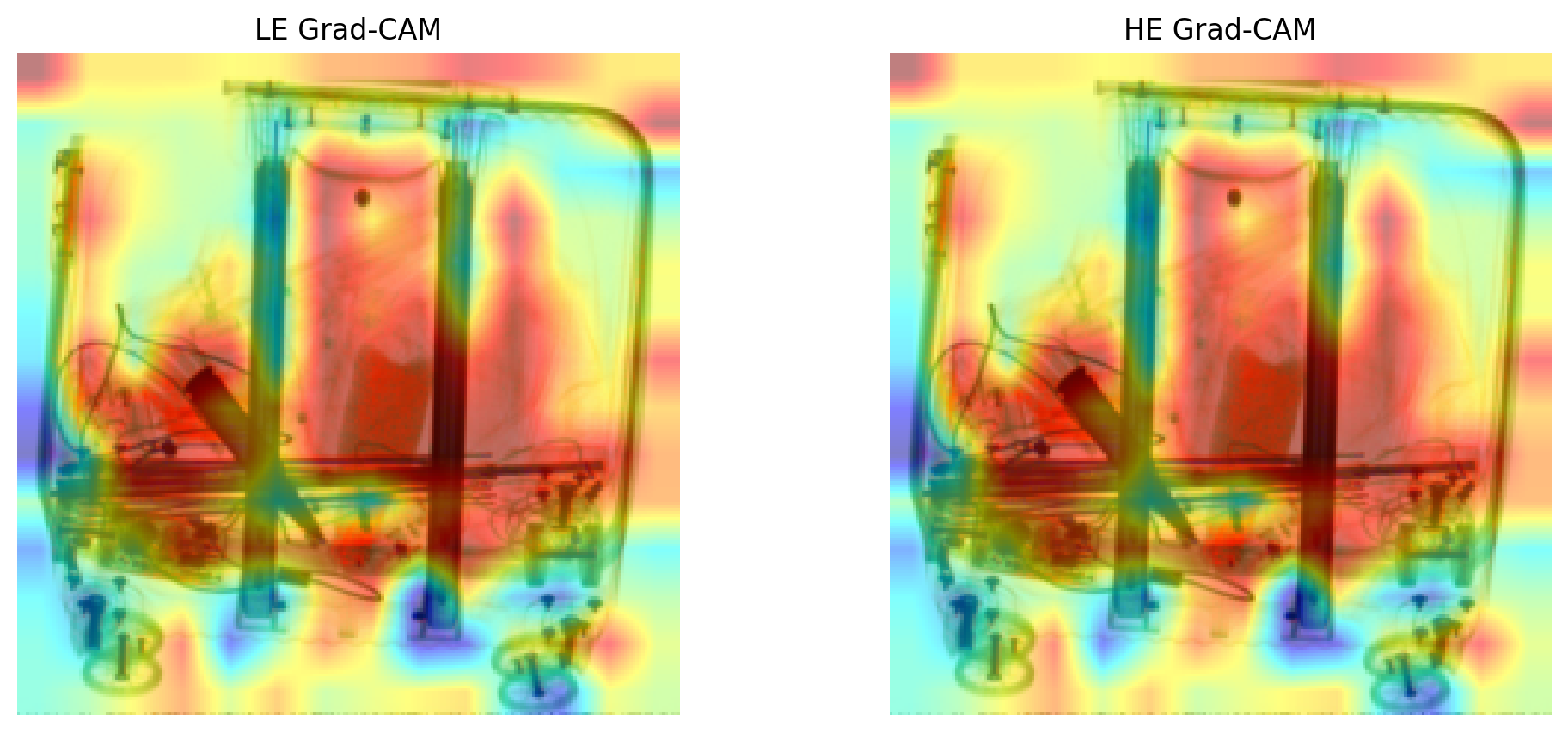}
    \caption{Branch-specific Grad-CAM (LE/HE).}
    \label{fig:gradcam_le_he}
  \end{subfigure}\hfill
  \begin{subfigure}[b]{0.48\linewidth}
    \includegraphics[width=\linewidth]{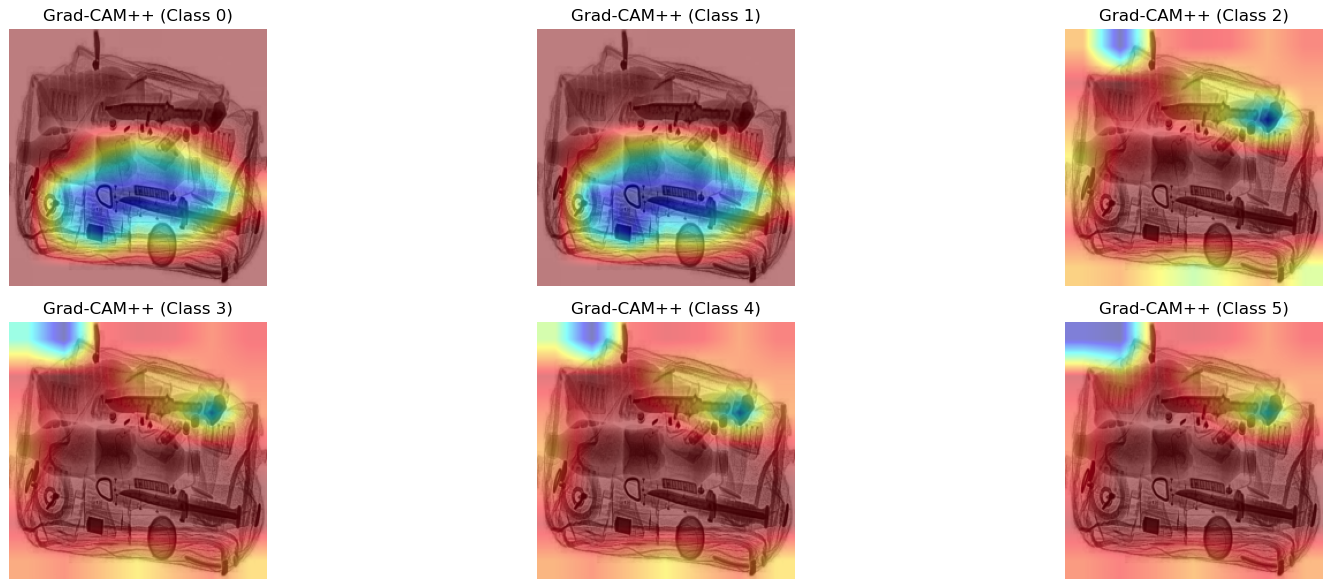}
    \caption{Class-conditioned Grad-CAM++.}
    \label{fig:gradcampp_classes}
  \end{subfigure}

  \begin{subfigure}[b]{0.48\linewidth}
    \includegraphics[width=\linewidth]{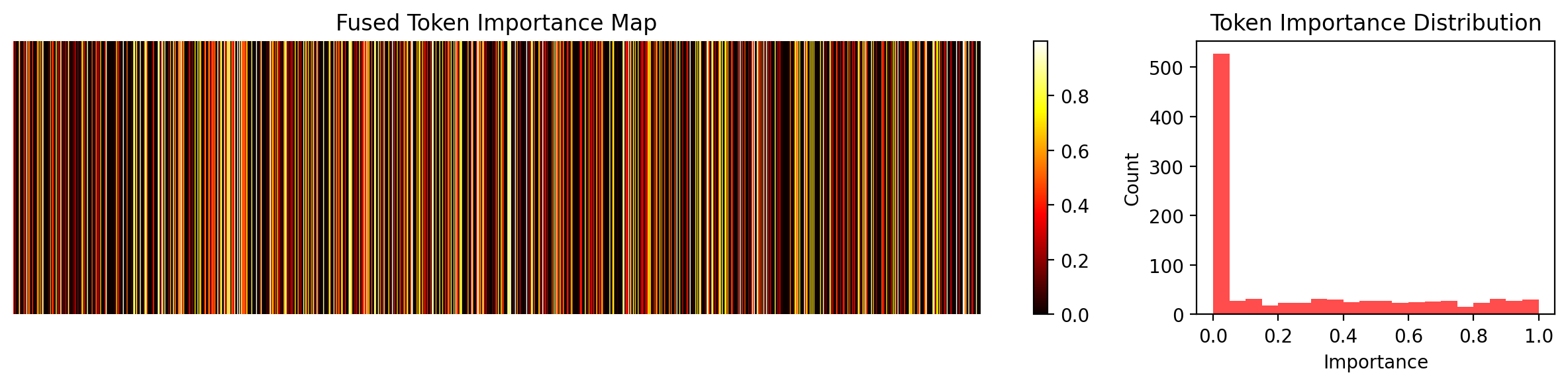}
    \caption{Fused token importance map and histogram.}
    \label{fig:fused_token}
  \end{subfigure}\hfill
  \begin{subfigure}[b]{0.48\linewidth}
    \includegraphics[width=\linewidth]{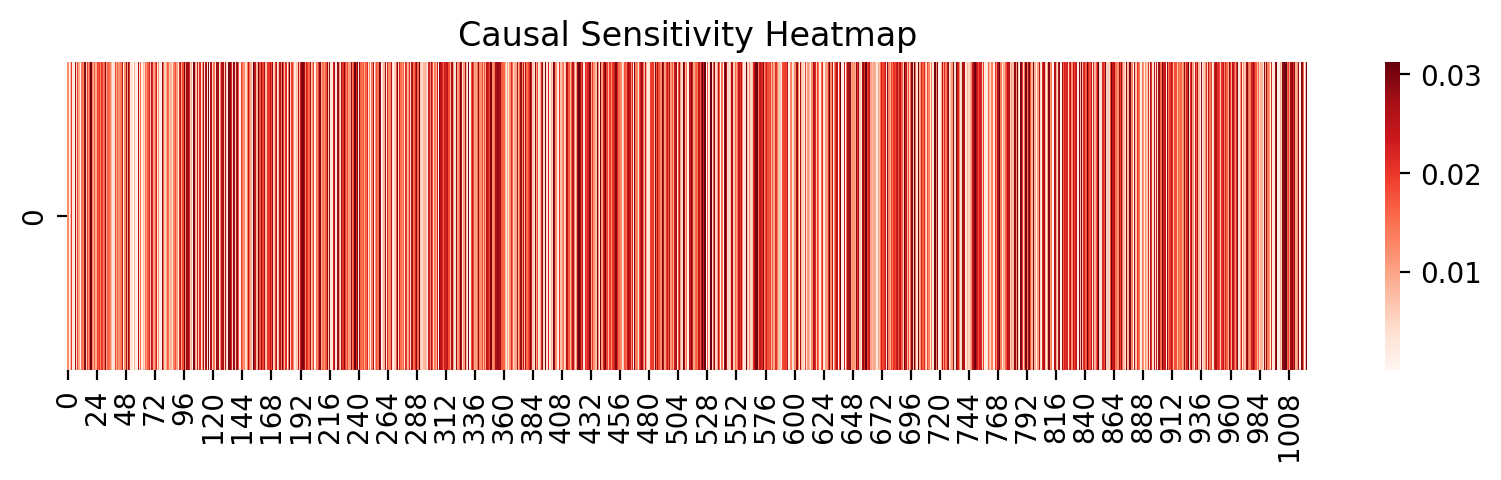}
    \caption{Causal sensitivity heatmap $\Delta_k$.}
    \label{fig:causal_heatmap}
  \end{subfigure}

  \caption{Interpretability analysis: (a) branch-specific Grad-CAM, (b) class-conditioned Grad-CAM++, (c) token importance map and histogram, (d) causal sensitivity heatmap.}
  \label{fig:gradcam_le_he_all}
\end{figure}

LE activations span low-density organic regions; HE activations localise dense metallic structures. Token importance is sparse ($>500/1024$ with importance $<0.05$). $\Delta_{\max}{\approx}0.028$ (reduced from $0.030$ pre-CIR), confirming CIR reduces single-token dependence as predicted by Theorem~\ref{thm:main}. Class-conditioned Grad-CAM confirms class-specific fidelity: Gun on barrel/handle, Knife on blade silhouette, as illustrated in Fig.~\ref{fig:gradcam_le_he_all}.

% ================================================================
% SECTION 6 — CONCLUSION
% ================================================================
\section{CONCLUSION}
\label{sec:conclusion}
We presented CRaFT-XDet, a causality-grounded and physics-motivated framework for dual-energy X-ray screening (29.2\,M params, 89.5\,GFLOPs, 345\,FPS). Theorem~\ref{thm:main} formalises that CIR approximates $P(Y \mid \mathrm{do}(X_{\mathrm{obj}}))$; Corollary~\ref{cor:generalisation} bounds cross-domain degradation by background distribution distance. SIXray: $0.9389$ mAP@0.5, $0.8021$ mAP@0.5:0.95. Zero-shot: $0.8851$ (OPIXray), $0.8172$ (HiXray), surpassing tuned baselines. CIR delivers the largest gains on Knife ($+3.1$\,pp) and Lighter ($+2.7$\,pp). Binary: $93.92\%$ recall, $0.9031$ IoU. The pseudo-environment partition approximates rather than recovers true scanner metadata; the environment-count ablation bounds the finite gap at $\approx0.9$\,pp. The RGB decomposition (Eq.~\eqref{eq:rgb_split}) is a \emph{physics-motivated heuristic} requiring re-calibration for non-standard colorimetric pipelines. Theorem~\ref{thm:main} uses $\lambda_{\mathrm{irm}}{\to}\infty$, $|\mathcal{E}|{\to}\infty$ as idealisations; finite-environment theoretical bounds remain open. Future work will investigate video-rate screening and extension of CIR to medical imaging.

\section*{DATA AND CODE AVAILABILITY}

The datasets used in this study are publicly available from the SIXray~\cite{miao2019sixray}, OPIXray~\cite{wei2020occluded}, and HiXray~\cite{wang2021hixray} repositories under their respective licensing terms. To promote reproducibility, the complete source code, trained model weights, data preprocessing pipeline, training and evaluation scripts, and instructions for reproducing the reported experiments will be made publicly available through a public GitHub repository upon publication of this article.
% \section*{Acknowledgements}

% \vfill\pagebreak
\bibliographystyle{IEEEtran}
\bibliography{bibliography}

\end{document}